%% file: neurips_2025.tex
\documentclass{article}

\PassOptionsToPackage{numbers, compress}{natbib}

 \usepackage[dblblindworkshop, final]{neurips_2025}

\usepackage[utf8]{inputenc} 
\usepackage[T1]{fontenc}
\usepackage{url}
\usepackage{booktabs}
\usepackage{amsfonts}
\usepackage{nicefrac}
\usepackage{xcolor}
\usepackage{soul}
\usepackage{amsmath}
\usepackage{multirow}
\usepackage{tikz}
\usetikzlibrary{shapes,arrows,positioning,decorations.pathreplacing,fit,backgrounds}
\usepackage{float}

\title{CAT: Curvature-Adaptive Transformers \\ for Geometry-Aware Learning}
\workshoptitle{Non-Euclidean Foundation Models and Geometric Learning}

\author{ 
    Ryan Y. ~Lin \quad Siddhartha ~Ojha \quad Nicholas Z. ~Bai
    \\
    Division of Engineering and Applied Science\\
    California Institute of Technology\\
    Pasadena, CA 91125, USA \\
    \texttt{\{rylin, sojha, nbai\}@caltech.edu} \\
}

\begin{document}

\maketitle

\begin{abstract}
    Transformers achieve strong performance across diverse domains but implicitly assume Euclidean geometry in their attention mechanisms, limiting their effectiveness on data with non-Euclidean structure. While recent extensions to hyperbolic and spherical spaces show promise for hierarchical and cyclical patterns, respectively, they require committing to a single geometry a priori, reducing flexibility when data exhibits mixed geometric properties. We introduce the Curvature-Adaptive Transformer (CAT), a novel architecture that dynamically learns per-token routing across three geometric attention branches through a lightweight, differentiable gating mechanism. Unlike fixed-geometry approaches, CAT enables adaptive geometric specialization, routing tokens to the appropriate curvature based on their local relational structure. The routing network provides interpretable curvature preferences while each branch employs geometry-specific operations optimized for its respective manifold. On knowledge graph completion benchmarks (FB15k-237, WN18RR), CAT achieves approximately 10\% improvements in MRR and Hits@10 over fixed-geometry baselines with minimal overhead (5\% parameter increase, comparable inference time). These results demonstrate that learned geometric adaptation outperforms any single fixed geometry for complex relational reasoning, establishing CAT as a scalable and interpretable foundation for mixture-of-geometry architectures across language, vision, and multimodal domains.
\end{abstract}

\section{Introduction}
Transformers have revolutionized machine learning across domains, including language~\citep{devlin2018bert,brown2020language}, vision~\citep{dosovitskiy2020image}, and scientific computing~\citep{jumper2021highly}. Their success stems from self-attention's ability to model flexible relationships through learned query-key interactions. However, transformers implicitly assume Euclidean geometry, treating tokens as points in flat space, an assumption that may be fundamentally limiting for data with inherent non-Euclidean structure.

Real-world data often exhibits geometric properties better captured in curved spaces. Hierarchical relationships in knowledge graphs and taxonomies naturally align with hyperbolic geometry's exponential volume growth~\citep{nickel2017poincareembeddingslearninghierarchical}, while cyclical patterns in temporal sequences and periodic phenomena are well-suited to spherical geometries~\citep{cohen2018sphericalcnns}. Recent geometric deep learning advances demonstrate that matching model geometry to data structure yields substantial performance gains~\citep{yang2024hypformerexploringefficienthyperbolic,cho2022sphericaltransformer}.

However, existing geometric transformers require committing to a single geometry a priori, failing to capture the heterogeneous geometric properties that characterize real datasets. Consider knowledge graphs, which simultaneously contain hierarchical taxonomic relations (hyperbolic-suited), symmetric equivalence relations (Euclidean-suited), and cyclical temporal patterns (spherical-suited). Fixed-geometry approaches cannot effectively model this complexity.

To address this fundamental limitation, we introduce the Curvature-Adaptive Transformer (CAT), which addresses this through dynamic, token-level geometry selection. CAT extends transformers with a differentiable routing mechanism that learns to assign each token to the most appropriate geometric attention branch, Euclidean, hyperbolic, or spherical, within a single forward pass. This eliminates a priori geometric assumptions while maintaining computational efficiency. Our key insight is that geometric specialization should be learned, not prescribed. By treating geometry selection as a routing problem inspired by mixture-of-experts~\citep{MoE}, CAT discovers and exploits heterogeneous geometric structure in complex datasets. Each branch employs principled manifold operations (Möbius transformations for hyperbolic, geodesic computations for spherical), ensuring mathematically consistent reasoning.

Our proposed model architecture carries four critical advantages: (1) Token-level adaptivity enabling fine-grained specialization, (2) General applicability to any sequential data, (3) Inherent interpretability through geometric routing weights, and (4) End-to-end optimization discovering complementary representations. To empirically validate these advantages, we evaluate CAT on knowledge graph completion (FB15k-237, WN18RR), achieving $\sim$10\% improvements in MRR and Hits@10 over best fixed-geometry baselines with minimal overhead (5\% parameter increase, comparable inference time). We argue that our novel approach represents a paradigm shift from "choosing the right geometry" to "learning to choose geometries dynamically," establishing foundations for geometric mixture-of-experts architectures across domains.

\section{Background \& Related Works}
\textbf{Transformers and Attention Mechanisms}
The transformer architecture, introduced by \citet{vaswani2017attention}, revolutionized sequence modeling through its self-attention mechanism, eliminating the need for recurrent or convolutional operations. Built on the principle of "attention is all you need," transformers compute attention weights between all pairs of positions in a sequence, enabling parallel computation and effective modeling of long-range dependencies. The core innovation lies in the scaled dot-product attention mechanism,
\[
\text{Attention}(Q,K,V) = \text{softmax}\left(\frac{QK^T}{\sqrt{d_k}}\right)V,
\]
where queries $Q$, keys $K$, and values $V$ are learned linear projections of the input embeddings. The flexibility to model dependencies has enabled transformer models to underpin SOTA advances in diverse domains. Yet, the computations underlying transformers implicitly subscribe to Euclidean geometry, which may not be optimal for some data~\citep{he2025positioneuclideanfoundation}, motivating the exploration of geometric extensions to the transformer architecture.

\textbf{Geometric Deep Learning and Non-Euclidean Transformers} While traditional neural networks operate in Euclidean space, implicitly assuming flat geometry with zero curvature, real-world data often exhibits intrinsic geometric structure that can be better captured in non-Euclidean spaces~\citep{he2025positioneuclideanfoundation}. Recent advances in geometric deep learning have empirically demonstrated that the choice of geometric space can fundamentally impact model expressiveness and performance~\citep{shen2023moleculargeometricdeeplearning, Pineda_2023, garcíavinuesa2025geometricdeeplearningassists}.

Characterized by negative curvature, hyperbolic geometry naturally represents hierarchal structures due to its exponential volume growth. Early work introduced hyperbolic embeddings for hierarchical data by embedding data into an $n$-dimensional Poincar\'e ball~\citep{nickel2017poincareembeddingslearninghierarchical}, followed by the introduction of hyperbolic neural networks~\citep{ganea2018hyperbolicentailmentconeslearning}. More recently, Hypformer extended transformers to hyperbolic space~\citep{yang2024hypformerexploringefficienthyperbolic}, demonstrating improved performance on hierarchical tasks. However, 
a critical limitation of these early methods in that they assume \emph{a priori} that data exhibits hierarchical structure and commit to hyperbolic geometry throughout the entire model in hopes of exploiting such structure.

Spherical geometry, with positive curvature, excels at representing cyclical patterns and angular relationships. Spherical CNN~\citep{cohen2018sphericalcnns} and subsequent transformer extensions~\citep{lai2023sphericaltransformerlidarbased3d, cho2022sphericaltransformer} have shown success in domains with natural spherical structure, such as 360\textdegree  images\citep{eder2020tangentimagesmitigatingspherical} and climate modeling~\citep{bonev2023sphericalfourierneuraloperators}.  Like hyperbolic approaches, these methods assume uniform geometric structure across the data, and in turn, commit to exploiting that geometric structure beforehand.

\textbf{Mixed-Curvature and Adaptive Geometry Approaches}
Recognizing the limitations of committing to a single geometry beforehand, recent work has begun exploring mixed-curvature spaces to handle data with heterogeneous geometric properties. Product manifolds~\citep{gu2018learning, skopek2020mixedcurvaturevariationalautoencoders} combine multiple geometric spaces, but require manual specification of the manifold structure and are limited to embedding and simple generative tasks.

Other works propose learning and adaptively selecting the appropriate geometry for a given setting, specifically in the context of augmenting graph convolutional networks (GCNs) or graph transformers~\citep{bachmann2020constantcurvaturegraphconvolutional, cho2023curveattentionmixedcurvaturetransformers}. \citet{cho2023curveattentionmixedcurvaturetransformers}, for example, introduces a graph-only transformer architecture that learns a fixed curvature per layer on graph inputs, scaling attention geometrically. To summarize, these existing approaches are limited to graph-based data modalities and lack the flexibility for per-token or region-specific curvature adaptation. 

\textbf{Mixture-of-Experts (MoE) and Routing Mechanisms}
A powerful paradigm for scaling capabilities of neural network while maintaining computational efficiency, mixture-of-experts (MoE) architectures, originally emerging in the 1990s, hinge on the simple idea of routing inputs to specialized expert networks based on learned gating functions, enabling conditional computation and expert specialization~\citep{MoE, MoE_survey}. Previous work has explored leveraging a MoE-type architectures to place graphs in mixture curvature spaces and shows promise as a direction for future work in graph foundation models~\citep{guo2024graphmoremitigatingtopologicalheterogeneity}.

In the traditional MoE architecture, the gating function learns to route to discrete experts. However, recent work has built upon this to develop  routing that allows gradient flow through multiple expert branches simultaneously~\citep{muqeeth2024softmergingexpertsadaptive}. We take this approach to mixing experts to allow the model to learn smooth interpolations between different geometric spaces while simultaneously avoiding the discrete optimization challenges associated with hard routing~\citep{bengio2013estimatingpropagatinggradientsstochastic}.

A limitation of classical MoE architectures, much like many other deep learning approachs, is in understanding the learned gating function and experts~\citep{chen2022understandingmixtureexpertsdeep, ismail2023interpretablemixtureexperts}. By designing experts each specialized to a specific geometry, we reap inherent interpretability, as the routing weights directly indicate the geometric properties the model has learned for each token.

\textbf{Our Contributions} Recognizing these limitations of existing learning approaches that seek to leverage non-Euclidean geometry, we introduce the Curvature-Adaptive Transformer (CAT), which addresses these highlighted limitations in the following ways:

\begin{itemize}
    \item \textbf{Token-Level Dynamic Geometry Selection} Unlike fixed-geometry approaches, CAT learns to route each token through the most appropriate geometric space based on learned structural preferences, eliminating the need for \emph{a priori} geometric assumptions, or even educated guesses. The per-token approach allows for richer granularity in geometry switching. 
    \item \textbf{General-Purpose Architecture} We provide the first \emph{general-purpose} transformer architecture that integrates Euclidean, hyperbolic, and spherical geometries within standard attention mechanisms, applicable to any sequential data, such as language, vision, time series, or graph-like data.
    \item \textbf{Interpretable Routing} Our routing mechanism is grounded in geometric principles, with each block corresponding to a specific geometry rather than arbitrary parameter specialization, providing clear interpretable insights into the geometric structure of sequential data.
    \item \textbf{End-to-End Optimization} All geometric branches and routing weights are jointly optimized, allowing the model to learn complementary geometric representations in a unified training loop rather than simply ensembling independent models.
\end{itemize}

Our work represents a fundamental shift from "choosing the right geometry" to "learning to choose (potentially mixed) geometries dynamically," opening new research directions in adaptive neural architectures and interpretable geometric representation learning.

\newpage
\section{Curvature-Adaptive Transformer (CAT) Architecture}

\input{figs/arch_fig}

We introduce the \textbf{Curvature-Adaptive Transformer (CAT)}, a modular attention mechanism that dynamically routes token representations through one of three geometry-specific attention branches: Euclidean, hyperbolic, or spherical. Loosely inspired by the manifold-based operations outlined in \cite{he2025positioneuclideanfoundation}, CAT enables continuous adaptation to the local relational structure of input tokens by learning per-token curvature preferences.

Given an input sequence \(\mathbf{X} \in \mathbb{R}^{B \times N \times d}\), where \(B\) is the batch size, \(N\) is the number of tokens, and \(d\) is the embedding dimension,   computes geometry-specific attention outputs \(\mathbf{Y}^{(g)}\) for \(g \in \{E, H, S\}\) (Euclidean, Hyperbolic, Spherical), and combines them via a learned routing distribution:
\[
\boldsymbol{\alpha} = \mathrm{softmax}(\mathrm{MLP}(\mathbf{X})) \in \mathbb{R}^{B \times N \times 3},
\]
\[
\mathbf{Y}_{b,n,:} = \sum_{g \in \{E,H,S\}} \alpha_{b,n,g} \, \mathbf{Y}^{(g)}_{b,n,:},
\]
where \(\alpha_{b,n,g}\) denotes the routing weight for geometry \(g\) at token \(n\) in batch \(b\). Each geometry-specific branch contains an attention mechanism and feedforward network, described below.

\subsection{Euclidean Attention}

The Euclidean branch uses standard Transformer components:
\begin{align*}
\mathbf{Z}^{(E)} &= \mathrm{MultiHeadAttn}(\mathbf{X}), \\
\mathbf{H}^{(E)} &= \mathrm{LayerNorm}(\mathbf{X} + \mathbf{Z}^{(E)}), \\
\mathbf{Y}^{(E)} &= \mathrm{LayerNorm}(\mathbf{H}^{(E)} + \mathrm{FF}(\mathbf{H}^{(E)})).
\end{align*}

\subsection{Hyperbolic Attention (Poincaré Ball)}

We use the Poincaré ball model \(\mathbb{B}_c = \{\mathbf{x} \in \mathbb{R}^d : \|\mathbf{x}\| < 1/\sqrt{c}\}\), with constant negative curvature \(-c\). Input tokens are mapped to the manifold via the exponential map at the origin:
\begin{align*}
\mathbf{q}_H &= \mathrm{proj}_{\mathbb{B}_c}(\exp_0(\mathbf{W}_q \mathbf{X})), \\
\mathbf{v}_H &= \mathrm{proj}_{\mathbb{B}_c}(\exp_0(\mathbf{W}_v \mathbf{X})),
\end{align*}
where \(\exp_0\) maps Euclidean vectors from the tangent space at the origin to the manifold, and \(\mathrm{proj}_{\mathbb{B}_c}\) ensures numerical stability. Pairwise attention weights are computed via hyperbolic distances:
\[
A_{ij} = \frac{\exp(-d_{\mathbb{B}_c}(\mathbf{q}_{H,i}, \mathbf{q}_{H,j}))}{\sum_k \exp(-d_{\mathbb{B}_c}(\mathbf{q}_{H,i}, \mathbf{q}_{H,k}))},
\]
followed by Möbius-weighted aggregation and logarithmic projection back to Euclidean space:
\[
\mathbf{Y}^{(H)} = \mathrm{FF}\left(\log_0\left(\mathrm{proj}_{\mathbb{B}_c}\left(\sum_j A_{ij} \odot \mathbf{v}_{H,j} \right)\right)\right),
\]
where \(\odot\) denotes Möbius scalar multiplication, and \(\log_0\) is the logarithmic map at the origin.

\subsection{Spherical Attention}

For positive curvature, we embed tokens on the unit hypersphere \(\mathbb{S}^{d} = \{ \mathbf{x} \in \mathbb{R}^{d+1} : \|\mathbf{x}\| = 1 \}\). Each token is lifted to \(\mathbb{R}^{d+1}\) and mapped onto the manifold via the exponential map at \(\mu = [\mathbf{0}, 1] \in \mathbb{S}^{d+1}\):
\[
\tilde{\mathbf{X}} = \exp_{\mu}([\mathbf{X}; \mathbf{0}]).
\]
Attention weights are computed via spherical similarity (cosine similarity on the manifold):
\[
A_{ij} = \frac{\exp(\langle \tilde{\mathbf{x}}_i, \tilde{\mathbf{x}}_j \rangle)}{\sum_k \exp(\langle \tilde{\mathbf{x}}_i, \tilde{\mathbf{x}}_k \rangle)},
\]
and outputs are projected back via the logarithmic map and a weighted aggregation:
\[
\mathbf{Y}^{(S)} = \mathrm{FF}\left(\log_{\mu}\left(\sum_j A_{ij} \cdot \tilde{\mathbf{x}}_j\right)\right).
\]

\subsection{Geometry Mixing}

The outputs \(\mathbf{Y}^{(E)}, \mathbf{Y}^{(H)}, \mathbf{Y}^{(S)}\) are combined through a token-wise convex mixture using the learned routing weights:
\[
\mathbf{Y} = \alpha_E \odot \mathbf{Y}^{(E)} + \alpha_H \odot \mathbf{Y}^{(H)} + \alpha_S \odot \mathbf{Y}^{(S)},
\]
where \(\alpha_g = \boldsymbol{\alpha}_{[..., g]} \in \mathbb{R}^{B \times N \times 1}\) and broadcasting is applied over the feature dimension. 

\subsection{Routing Mechanism: Motivation and Benefits}

The CAT introduces a mechanism for dynamic geometric specialization at the token level by learning \textit{curvature-aware routing weights}. Instead of committing to a single global geometry, CAT enables each token to softly select among three attention branches, Euclidean, hyperbolic, and spherical, allowing the model to adapt its inductive bias based on the local relational structure of the input.

This design is motivated by the observation in prior works that different relational structures are suited for distinct data topologies ~\citep{nickel2017poincareembeddingslearninghierarchical, gu2018learning}: Euclidean space effectively models flat or grid-like layouts; hyperbolic space naturally captures tree-like or hierarchical structures due to its exponential volume growth; and spherical space is ideal for cyclic or angular relationships, as seen in periodic sequences or directional data. In real-world tasks with heterogeneous or multi-scale dependencies, a fixed geometric inductive bias can be limiting. CAT addresses this by learning to interpolate geometries in a data-driven manner.

In CAT, all geometry-specific operations are performed intrinsically within their respective manifolds. Each branch then maps its outputs back to Euclidean space through the logarithmic map, \emph{before} entering any mixing stages. This ensures that intra-branch computations remain geometrically sound, while the cross-branch combination is well-defined and stable in a shared space. Although this approach relaxes strict manifold consistency at the point of mixture, it offers a principled and efficient mechanism for adaptive geometry selection.

Moreover, all operations involved, including exponential and logarithmic maps on the manifolds, curvature-dependent distance calculations, and attention mechanisms, are differentiable. This makes CAT fully compatible with gradient-based training and backpropagation. Additionally, because all three geometric attentions are computed \textit{in parallel} within the same forward pass, CAT achieves curvature adaptivity without significant runtime penalties.

Importantly, this multi-geometry design incurs only a modest parameter overhead. The only additional learnable component is the selector MLP, which is lightweight relative to the rest of the model. This efficiency allows CAT to scale to large architectures without prohibitive cost.

Empirically, the learned routing mechanism enables the model to identify and exploit the most appropriate geometric representation for each token, improving generalization in tasks that feature complex or mixed relational patterns. Furthermore, the per-token routing weights provide \emph{interpretability}: they offer insight into the geometric assumptions the model leverages across different inputs, effectively revealing how curvature preferences vary by context.

\newpage
\section{Experiments}
\label{sec:experiments}
We evaluate CAT in a controlled, low-parameter setting to assess its geometric adaptability under tight capacity constraints. Our primary goal is not to achieve state-of-the-art performance, but rather to demonstrate that CAT provides meaningful improvements over fixed-geometry alternatives. In other words, we value relative performance gains in our experiments as opposed to absolute numbers. Nonetheless, we do find that CAT delivers competitive results.

Specifically, we target the link prediction task on knowledge graph datasets as a proxy for more complex relational reasoning problems. This setup enables clear comparisons across models with equivalent architecture and parameter budgets.

Our models contain approximately 1M parameters for FB15k-237 and 2.7M parameters for WN18RR, which aligns with other lightweight Transformer-style models reported in the literature, such as TransE, DistMult, and RotatE~\citep{NIPS2013_1cecc7a7, yang2015embeddingentitiesrelationslearning, sun2018rotate}. Even in this compact setting, CAT delivers competitive performance relative to its parameter count. For instance, on FB15k-237, it achieves comparable Hits@10 to KG-BERT~\citep{yao2019kgbertbertknowledgegraph}, a model with roughly 100M parameters, while surpassing it in MRR, demonstrating that our evaluation reflects meaningful results rather than a mere toy example (KG-BERT has a Hits@10 of 0.524 and MRR of 0.216, whereas CAT achieves 0.473 and 0.290 in the same metrics, respectively).

By focusing on compact, single-block architectures, we isolate the effect of curvature adaptivity from confounding factors such as depth, parameter scaling, or clever training tricks. This small-scale setting allows us to ask: \emph{Does a token-level curvature-adaptive attention mechanism provide tangible benefits over fixed-geometry models at this scale?} Positive results here would support the idea that adaptive geometry is a meaningful inductive bias, even in low-capacity models, making it a strong candidate for scaling to more complex architectures and downstream tasks such as classification/processing, retrieval, and multi-hop reasoning. We evaluate performance on two standard knowledge graph completion benchmarks, \textbf{FB15k-237}~\citep{NIPS2013_1cecc7a7} and \textbf{WN18RR}~\citep{dettmers2018conve}, comparing CAT to matched, fixed-geometry baselines that operate solely in Euclidean, hyperbolic, or spherical attention spaces.

\subsection{Link Prediction Setup}

In the link prediction task, the model is given a partial triple \((h, r, ?)\) and must score all possible tail entities. We adopt a standard scoring architecture used across all baselines:

\begin{itemize}\setlength\itemsep{0pt}
    \item \textbf{Embeddings}: Each entity and relation is represented by a learnable vector in \(\mathbb{R}^d\), composed as \(x = \mathrm{Drop}(h + r)\).
    \item \textbf{Transformer block}: The composed vector is passed through a geometry-specific Transformer block (CAT or fixed-geometry variant).
    \item \textbf{Scoring}: The output is scored against all entity embeddings via dot product, producing logits for ranking.
\end{itemize}

All models share the same embedding dimensionality, dropout configuration, and optimization hyperparameters. The \emph{only} architectural difference lies in the use of adaptive versus fixed-curvature attention mechanisms.

\subsection{Baselines}

We compare the full CAT model against three geometry-locked variants:

\begin{itemize}\setlength\itemsep{0pt}
    \item \textbf{CAT (ours)}: Learns per-token routing weights over three geometry-specific attention branches (Euclidean, hyperbolic, and spherical).
    \item \textbf{Fixed-Euclidean}: Replaces CAT with standard Transformer attention operating solely in Euclidean space.
    \item \textbf{Fixed-Hyperbolic}: Uses only hyperbolic attention with Poincaré ball geometry.
    \item \textbf{Fixed-Spherical}: Employs spherical attention via cosine similarity on the unit hypersphere.
\end{itemize}

All fixed-geometry baselines are implemented using the same architecture and composition functions as CAT, differing only in their geometry-specific attention logic.

\subsection{Training Details}

All models are trained using standard cross-entropy loss over the entity vocabulary. Given a training triple \((h, r, t)\), where \(h, t \in \mathcal{E}\) (entities) and \(r \in \mathcal{R}\) (relations), the model produces a score vector \(\mathbf{s} \in \mathbb{R}^{|\mathcal{E}|}\) over all candidate tail entities. The predicted distribution is compared to the true tail entity \(t\) via a smoothed cross-entropy objective:
\[
\mathcal{L}_{\text{ce}} = - \sum_{i=1}^{|\mathcal{E}|} y_i \log p_i, \quad \text{where } \mathbf{p} = \mathrm{softmax}(\mathbf{s}) \text{ and } y_i = 
\begin{cases}
1 - \varepsilon & \text{if } i = t \\
\frac{\varepsilon}{|\mathcal{E}| - 1} & \text{otherwise}
\end{cases}
\]
with \(\varepsilon = 0.1\) representing the label smoothing coefficient.

In models using the CAT, we add an auxiliary entropy regularization term that encourages high-entropy routing distributions \(\boldsymbol{\alpha} \in \mathbb{R}^{B \times N \times 3}\), promoting usage of multiple geometries. This regularizer is given by:
\[
\mathcal{L}_{\text{entropy}} = \frac{1}{B N} \sum_{b=1}^B \sum_{n=1}^N \sum_{g \in \{E, H, S\}} -\alpha_{b,n,g} \log \alpha_{b,n,g}
\]
and the final loss becomes $\mathcal{L} = \mathcal{L}_{\text{ce}} + \lambda_{\text{ent}} \cdot \mathcal{L}_{\text{entropy}},$ where \(\lambda_{\text{ent}}\) is a weight annealed over time.

The entropy regularization is motivated by the desire to prevent premature collapse of the routing distribution to a single geometry, which could hinder the model's ability to flexibly exploit the diverse geometric inductive biases during early stages of training. By encouraging higher entropy, the model is nudged to explore and utilize multiple geometric representations, fostering richer and more adaptive feature learning. Over time, annealing \(\lambda_{\text{ent}}\) allows the model to sharpen the routing as appropriate.

For evaluation, we follow the standard filtered ranking protocol: each test triple \((h, r, t)\) is scored against all candidate tails \(t' \in \mathcal{E}\), and known true triples \((h, r, t') \in \mathcal{T}_{\text{train}} \cup \mathcal{T}_{\text{valid}} \cup \mathcal{T}_{\text{test}}\) are masked during ranking. We report Mean Reciprocal Rank (MRR) and Hits@10 as evaluation metrics. Full hyperparameter settings, optimizer details, and other training settings are deferred to Appendix~\ref{appendix:training-parameters}.

\subsection{Results on Knowledge Graph Completion}\label{sec:results}
\begin{table}[ht!]
\caption{Average link prediction performance on FB15k-237 and WN18RR over 50 runs, $\pm$ 1 standard deviation. CAT consistently outperforms fixed-geometry baselines on both MRR and Hits@10 metrics. All models are trained to convergence using identical optimizers, schedulers, and loss functions, with the only modification being the addition of a routing entropy term for CAT. }
\label{tab:link-pred-results}
\centering
\begin{tabular}{lccccc}
\toprule
\multirow{2}{*}{\textbf{Model}} & \multicolumn{2}{c}{\textbf{FB15k-237}} & \multicolumn{2}{c}{\textbf{WN18RR}}\\
\cmidrule(lr){2-3} \cmidrule(lr){4-5}
 & MRR & Hits@10 & MRR & Hits@10 \\
\midrule
Fixed-Euclidean & 0.2706 $\pm$ 0.0010 & 0.4486 $\pm$ 0.0011 & 0.2161 $\pm$ 0.0037 & 0.4553 $\pm$ 0.0050 \\
Fixed-Hyperbolic & 0.2655 $\pm$ 0.0014 & 0.4276 $\pm$ 0.0024 & 0.0918 $\pm$ 0.0051 & 0.1667 $\pm$ 0.0089 \\
Fixed-Spherical & 0.2576 $\pm$ 0.0008 & 0.4124 $\pm$ 0.0013 & 0.0800 $\pm$ 0.0097 & 0.1424 $\pm$ 0.0187 \\
\textbf{CAT (ours)} & \textbf{0.2904 $\pm$ 0.0007} & \textbf{0.4730 $\pm$ 0.0010} & \textbf{0.2417 $\pm$ 0.0028} & \textbf{0.5016 $\pm$ 0.0056} \\
\bottomrule
\end{tabular}
\end{table}

We evaluate link prediction performance using standard metrics: Mean Reciprocal Rank (MRR) and Hits@10. Table~\ref{tab:link-pred-results} reports average results over 50 runs on the FB15k-237 and WN18RR datasets. Across all metrics, our CAT consistently achieves a $\sim10\%$ relative improvement compared to the best-performing fixed-geometry transformer for each metric/dataset while maintaining comparable model capacity (see Table~\ref{tab:params-runtime}), demonstrating the importance of the adaptive routing mechanism.
Examining the WN18RR dataset in particular, we observe that transformers with fixed hyperbolic or spherical geometries perform substantially worse than their Euclidean counterpart. This disparity aligns with the geometric characteristics of WN18RR, which contains a heterogeneous mixture of hierarchical, symmetric, and compositional relations. Such diversity leads to a relational structure better approximated by a "flatter" or more flexible geometry, which Euclidean space naturally provides. Importantly, our CAT not only matches but exceeds the performance of the Euclidean baseline on WN18RR. We attribute this gain to CAT’s ability to dynamically combine multiple geometric spaces, effectively leveraging Euclidean flexibility alongside the inductive biases of hyperbolic and spherical spaces. This enriched representation capacity enables CAT to model the complex and varied relational patterns present in WN18RR more effectively than any single fixed geometry.

\subsection{Parameter Efficiency and Runtime}\label{sec:efficiency}
To assess the efficiency of CAT, we benchmark each model’s parameter count and per-batch inference time on the FB15k-237 dataset using an NVIDIA A10 (24~GB) GPU. Parameter counts are reported for the full architecture, while inference times are measured with random weights to isolate architectural effects. For runtime evaluation, we use a batch size of 512 and report steady-state latencies averaged over 100 forward passes after 50 warmup runs, excluding one-time initialization costs. Results are shown in Table~\ref{tab:params-runtime}.

By sharing large components such as the embedding and feed-forward sublayers, CAT incurs minor overhead to fixed-geometry models despite evaluating three attention branches. Notably, CAT remains faster than naively combining separate fixed-geometry models, which would require 3x the parameters and significantly more computation.

\begin{table}[ht!]
\caption{Parameter and runtime comparison across models for both datasets. CAT achieves curvature adaptivity with only minor additional overhead. }
\label{tab:params-runtime}
\centering
\begin{tabular}{lccccc}
\toprule
\multirow{2}{*}{\textbf{Model}} & \multicolumn{2}{c}{\textbf{FB15k-237}} & \multicolumn{2}{c}{\textbf{WN18RR}}\\
\cmidrule(lr){2-3} \cmidrule(lr){4-5}
 & \# Parameters & Inference Time (ms) & \# Parameters & Inference Time (ms) \\
\midrule
Fixed-Euclidean & 979,264 &  0.781 & 2,654,528 & 1.012 \\
Fixed-Hyperbolic & 974,720 & 3.280 & 2,649,984 & 3.592 \\
Fixed-Spherical & 967,090 & 1.313 & 2,654,528 & 1.533 \\
\textbf{CAT (ours)} & 1,031,669 & 4.885 & 2,706,933 & 5.566 \\
\bottomrule
\end{tabular}
\end{table}

\vspace{0.5em}

\subsection{Routing Analysis}
To understand how CAT adapts its geometry during training, we examine the behavior of the routing mechanism by analyzing the average routing weights $\boldsymbol{\alpha}$ assigned to each geometry during the last training epoch on a FB15k-237 run.

\begin{figure}[ht!]
\centering
\includegraphics[width=0.6\linewidth]{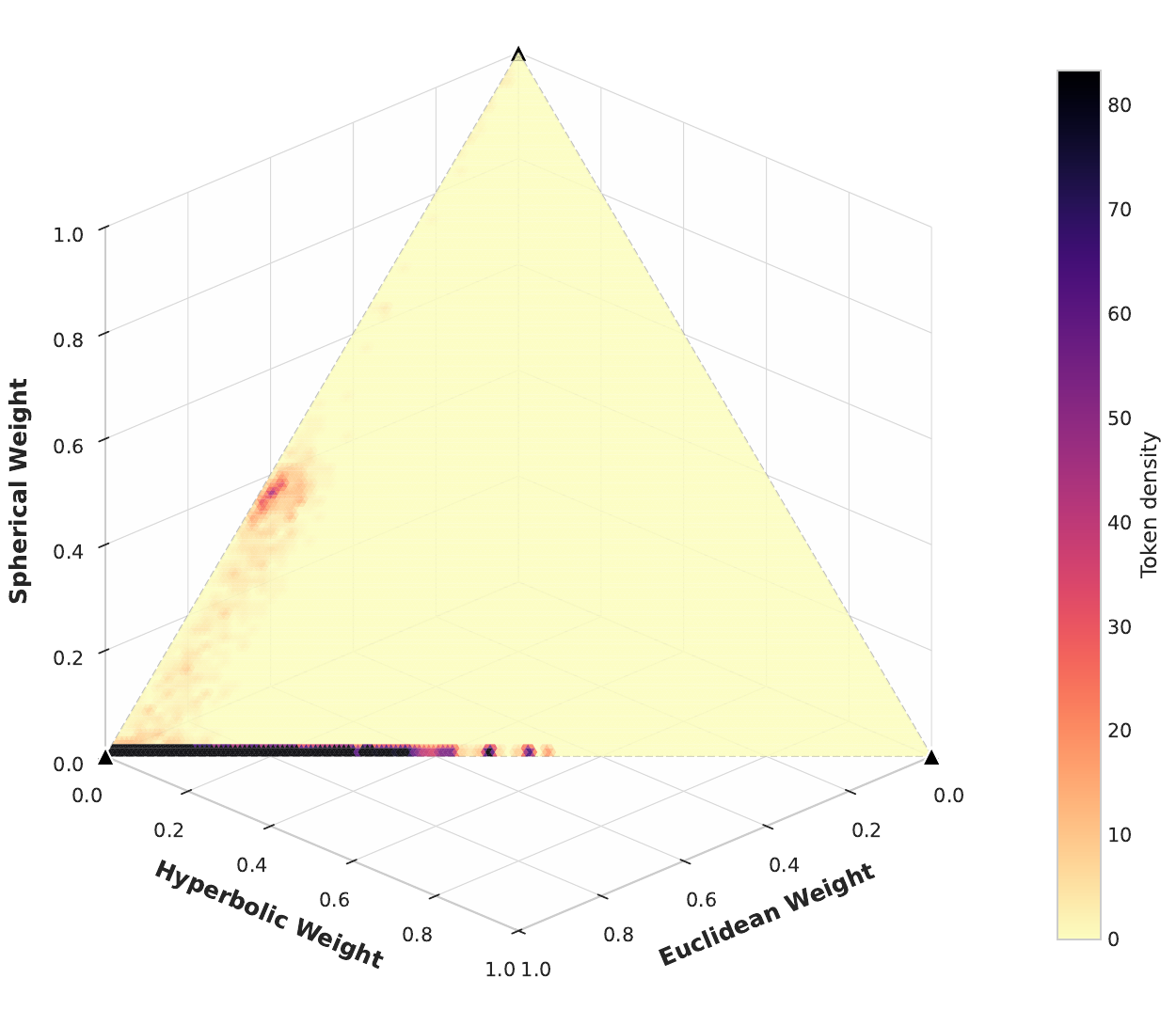}
\caption{Ternary heatmap visualizing routing weights across geometries on FB15k-237. Tokens concentrate towards Euclidean, with noticeable Euclidean–Hyperbolic mixtures. Spherical contributions remain negligible for this particular dataset.}
\label{fig:routing-density}
\end{figure}

Figure~\ref{fig:routing-density} shows that Euclidean space dominates overall, reflecting its strong alignment with local relational patterns. Hyperbolic weights are selectively activated for hierarchical structures, while spherical weights remain near zero with rare activations. To summarize this case, CAT leans towards Euclidean as the default backbone and applies geometry-specific corrections as needed. 

\section{Discussion and Conclusions}
\label{sec:discussion}
Our results demonstrate that curvature is not a one-size-fits-all inductive bias. CAT consistently outperforms similar capacity fixed-geometry baselines on FB15k-237 and WN18RR (Table~\ref{tab:link-pred-results}). Our analysis of routing weights (Fig.~\ref{fig:routing-density}) reveals that Euclidean attention serves as the default backbone for most tokens, while hyperbolic attention selectively engages for hierarchical structures, and spherical attention activates sparsely. As routing is differentiable and trained end-to-end, CAT discovers complementary niches across branches and mixes them optimally per token. The result is improved ranking quality with minimal overhead (Section \ref{sec:efficiency}), supporting the key idea that curvature should be meaningfully chosen by the model, not fixed a priori.

Interpretability follows directly from the design: experts are tied to explicit geometries. High hyperbolic mass is a useful signal of local tree-likeness; high spherical mass points to angular/cyclic patterns; high Euclidean mass reflects locally flat interactions. On our benchmarks, the low spherical usage should be read as a dataset property, not a general indictment of spherical attention; domains with periodicity or directional signals (e.g., time-of-day effects, 360$^\circ$ imagery) are natural candidates for higher spherical utilization.

Despite evaluating three branches per forward pass, CAT's overhead remains modest (See Section~\ref{sec:efficiency}) because the routing MLP is small and bulk parameters reside in shared embeddings and feed-forward layers. Entropy regularization during early training prevents premature collapse to Euclidean routing, encouraging exploration of non-Euclidean experts and leading to stronger final mixtures. Annealing this regularizer later allows the model to sharpen specializations, mirroring standard MoE practice. Our proposed CAT architecture dives deeper than layer-wise curvature selection to target structural heterogeneity at the local level, complementing previous attempts to better cater to potentially non-Euclidean geometries at a global level. 

Our evaluation focuses on two knowledge-graph benchmarks with compact models; broader studies across modalities (text, vision, time series) and scales are necessary to fully map where curvature adaptivity pays off. We note that such domains may be better suited to benefit from the spherical block, justifying training using all three blocks when deploying to new domains. Further, we have not yet explored conditional execution for runtime reduction, nor fine-grained attributions from routing to specific relation types. We view CAT as a foundation for granular mixture-of-geometry architectures: future work includes (i) dynamic per-layer geometry, (ii) branch pruning or distillation for fast inference, (iii) domain-specific case studies where spherical inductive biases are expected to dominate, and (iv) expanding CAT's flexible parallel-track architecture to incorporate even more geometries beyond the current three implemented.

We introduced the Curvature-Adaptive Transformer (CAT), a geometry-aware attention mechanism that learns per-token curvature selection. CAT achieves approximately 10\% relative improvement over fixed-geometry baselines with minimal overhead while providing interpretability through routing weights. Our findings support the key idea that geometry should be learned by the model, not fixed a priori. CAT establishes a foundation for scalable, interpretable mixture-of-geometry architectures, opening avenues for future exploration across language, vision, and multimodal domains.

\clearpage

\bibliographystyle{unsrtnat}
\bibliography{references}

\appendix

\section{Training Hyperparameters}\label{appendix:training-parameters}

This section provides a comprehensive overview of the training setup and hyperparameters used in all experiments. 

\paragraph{Model and Data Setup:}
\begin{itemize}
    \item \textbf{Embedding dimension:} \(d = 64\).
    \item \textbf{Datasets:} Experiments were conducted on FB15k-237 and WN18RR.
    \item \textbf{Batch size:} 512 samples per training iteration.
    \item \textbf{Entity and relation embeddings:} Learned via standard embedding layers initialized using Xavier uniform initialization \cite{glorot2010understanding}.
\end{itemize}

\paragraph{Optimization:}
\begin{itemize}
    \item \textbf{Optimizer:} AdamW \cite{loshchilov2019decoupled} with weight decay \(1 \times 10^{-3}\).
    \item \textbf{Initial learning rate:} 0.001.
    \item \textbf{Learning rate scheduler:} ReduceLROnPlateau with factor 0.5 and patience 10 epochs.
    \item \textbf{Number of epochs:} 200.
\end{itemize}

\paragraph{Regularization and Dropout:}
\begin{itemize}
    \item \textbf{Dropout rate:} 0.2 applied independently on entity embeddings, relation embeddings, and composite representations before passing to the Transformer block.
    \item \textbf{Label smoothing:} \(\varepsilon = 0.1\) used in cross-entropy loss to prevent overconfidence.
\end{itemize}

\paragraph{Entropy Regularization on Routing Weights:}
\begin{itemize}
    \item \textbf{Entropy regularization weight \(\lambda_{\text{ent}}\):} Initialized at 0.01.
    \item \textbf{Annealing schedule:} \(\lambda_{\text{ent}}\) is multiplied by 0.95 each epoch, with a minimum value capped at 0.001.
    \item \textbf{Purpose:} Encourages high-entropy routing distributions early in training to promote the use of multiple geometric attention pathways, avoiding premature collapse to a single geometry. The annealing schedule allows the model to gradually focus routing as training progresses.
\end{itemize}

\paragraph{Evaluation Protocol:}
\begin{itemize}
    \item \textbf{Filtered ranking evaluation} is used: for each test triple \((h, r, t)\), all candidate tail entities \(t' \in \mathcal{E}\) are scored, with any \((h, r, t')\) known to be true in training, validation, or test sets masked by setting their scores to \(-\infty\).
    \item \textbf{Metrics reported:} Mean Reciprocal Rank (MRR) and Hits@10.
\end{itemize}

\paragraph{Implementation Details:}
\begin{itemize}
    \item \textbf{Device:} Training performed on P100 GPUs with model and data tensors moved to the appropriate device.
    \item \textbf{Parameter counts:} All models have approximately 1-3 million trainable parameters, depending on the dataset, ensuring fair comparison at similar model capacity. Parameter counts and efficiency analysis is also included in Section \ref{sec:efficiency} of the main text.
    \item \textbf{Initialization:} Entity and relation embeddings are reinitialized at the start of each experiment using Xavier uniform initialization.
\end{itemize}

\paragraph{Benchmark Hardware \& Software Setup:} All benchmarks were run on a Lambda Cloud instance equipped with an NVIDIA A10 GPU (24 GB PCIe), 30 vCPUs, 200 GiB RAM, and 1.4 TiB SSD storage. We used the preinstalled Lambda Cloud PyTorch stack with CUDA and cuDNN support.

\paragraph{Code Availability:} The full training and evaluation code, including dataset preprocessing and model definitions, will be released upon publication.

\section{Node Classification on CORA}

We further evaluate CAT performance by evaluating on node classification tasks on CORA~\citep{cora_dataset}. The dataset consists of 2,708 publications classified into seven categories. Table \ref{tab:cora_results} shows our results across 10 runs. As with graph completion tasks in Section \ref{sec:results}, our method outperforms standard fixed-geometry transformers. 

\begin{table}[h]
    \caption{Comparison for node classification task on CORA. }
    \label{tab:cora_results}
    \centering
    \begin{tabular}{lcc}
    \toprule
    \textbf{Model} & \textbf{Train Accuracy} & \textbf{Test Accuracy} \\
    \midrule
    Fixed-Euclidean & $\mathbf{1.0000\pm 0.0000}$ & $0.5760 \pm 0.0061$ \\
    Fixed-Hyperbolic & $0.1429 \pm 0.0000$ & $0.2804 \pm 0.0779$ \\
    Fixed-Spherical & $0.1429 \pm 0.0000$ & $0.1759 \pm 0.0727$ \\
    CAT (ours) & $\mathbf{1.0000\pm 0.0000}$ &  $\mathbf{0.5811 \pm 0.0047}$ \\
    \bottomrule
    \end{tabular}
\end{table}

The results reveal several important limitations of both CAT and Fixed-Euclidean architecture. Notably, both methods appear to suffer from overfitting on small, homogeneous datasets such as CORA. We achieve near-perfect training accuracy, yet find only moderate test performance. The limited dataset size relative to model complexity prevents either architecture from learning generalizable representations. We also observe a substantial performance disparity between Fixed-Euclidean and non-Euclidean alternatives--Fixed-Hyperbolic ($0.2804$) and Fixed-Spherical ($0.1759$)--revealing that CORA naturally lends itself towards flat geometric structures. Despite this, CAT still attains accuracy exceeding fixed-geometry baselines. Results demonstrate that our approach effectively routes between geometric representations, even in homogeneous datasets. 

\section{Licenses For Assets Used}
\label{appendix:licenses}
The experiments in this work make use of several open-source libraries and datasets, all of which are cited and whose licenses are respected. \texttt{PyTorch}~\citep{paszke2019pytorchimperativestylehighperformance} is released under the BSD-3-Clause license. \texttt{PyTorch Geometric}~\citep{Fey_etal_2025} is distributed under the MIT license. \texttt{geoopt}~\citep{geoopt2020kochurov} is provided under the Apache-2.0 license.

\end{document}

%% file: figs/arch_fig.tex
\usetikzlibrary{fit,backgrounds} 

\begin{figure}[htbp]
\centering
\begin{tikzpicture}[
    node distance=0.7cm,
    box/.style={rectangle, rounded corners=3pt, minimum width=1.5cm, minimum height=0.8cm, align=center, draw=black, thick},
    branch/.style={rectangle, rounded corners=5pt, minimum width=1.5cm, minimum height=0.8cm, align=center, draw=black, very thick},
    euclidean/.style={branch, fill=green!25},
    hyperbolic/.style={branch, fill=blue!25},
    spherical/.style={branch, fill=red!25},
    routing/.style={box, fill=purple!25},
    mixing/.style={box, fill=gray!20},
    arrow/.style={->, thick, >=stealth},
    tiny/.style={font=\tiny}
]

\node[box, fill=gray!20] (input) {\textbf{Input} \\ \tiny $\mathbf{X}$};

\node[routing, right=of input] (routing) {\textbf{Routing} \\ \tiny $\boldsymbol{\alpha}$};

\node[hyperbolic, right=of routing, xshift=0.3cm] (hyperbolic) {\textbf{Hyperbolic} \\ \tiny $\mathbf{Y}^{(H)}$};
\node[euclidean, above=0.5cm of hyperbolic] (euclidean) {\textbf{Euclidean} \\ \tiny $\mathbf{Y}^{(E)}$};
\node[spherical, below=0.5cm of hyperbolic] (spherical) {\textbf{Spherical} \\ \tiny $\mathbf{Y}^{(S)}$};

\node[mixing, right=1cm of hyperbolic] (mixing) {\textbf{Mixing} \\ \tiny $\sum \alpha_g \mathbf{Y}^{(g)}$};

\node[box, fill=gray!20, right=of mixing] (output) {\textbf{Output} \\ \tiny $\mathbf{Y}$};

\draw[arrow] (input) -- (routing);
\draw[arrow] (routing.east) -- (euclidean.west);
\draw[arrow] (routing.east) -- (hyperbolic.west);
\draw[arrow] (routing.east) -- (spherical.west);
\draw[arrow] (euclidean.east) -- (mixing.north west);
\draw[arrow] (hyperbolic.east) -- (mixing.west);
\draw[arrow] (spherical.east) -- (mixing.south west);
\draw[arrow] (mixing) -- (output);

\node[above left=0.1cm of euclidean.west, tiny, green!70!black] {$\alpha_E$};
\node[above left=0.1cm of hyperbolic.west, tiny, blue!70!black] {$\alpha_H$};
\node[below left=0.1cm of spherical.west, tiny, red!70!black] {$\alpha_S$};

\begin{pgfonlayer}{background}
\node[
    draw=black!50,
    fill=gray!10,
    rounded corners,
    inner sep=0.3cm,
    fit=(routing) (euclidean) (hyperbolic) (spherical) (mixing)
] (catblockbox) {};
\node[anchor=north west, font=\scriptsize, xshift=0.1cm,yshift=0.5cm] 
at (catblockbox.south west) {CATBlock};
\end{pgfonlayer}

\end{tikzpicture}
\caption{CATBlock architecture: Input flows through routing MLP to three parallel geometry-specific branches, combined via learned \emph{per-token} weights.}
\label{fig:catblock_architecture}
\end{figure}